# Improved Local Search in Artificial Bee Colony using Golden Section Search

[1]Tarun Kumar Sharma, [2]Millie Pant, [3]V.P.Singh

[1, 2] Indian Institute of Technology Roorkee, Roorkee, India
[3] SCET, Saharanpur, India

Email:{taruniitr1; millidma, singhvp3}@gmail.com

**Abstract** –Artificial bee colony (ABC), an optimization algorithm is a recent addition to the family of population based search algorithm. ABC has taken its inspiration from the collective intelligent foraging behavior of honey bees. In this study we have incorporated golden section search mechanism in the structure of basic ABC to improve the global convergence and prevent to stick on a local solution. The proposed variant is termed as *ILS*-ABC. Comparative numerical results with the state-of-art algorithms show the performance of the proposal when applied to the set of unconstrained engineering design problems. The simulated results show that the proposed variant can be successfully applied to solve real life problems.

## 1. Introduction

Population based search heuristics like Genetic Algorithm (GA) [1], Particle Swarm Optimization (PSO) [2], Differential Evolution (DE) [3], Ant Colony Optimization (ACO) [4] and a like have attracted the researchers and scientist to solve complex problems arising in the domain of engineering design, finance, chemical etc. Population based search techniques are nontraditional search techniques which do not require any auxiliary properties like differentiability and continuity of the objective function as well as also independent of the nature of the problems. These algorithms can be successfully applied to solve many real life problems [5]-[7].

Artificial Bee Colony (ABC), proposed by Karaboga [8]-[10] is the recent addition to the population based search heuristics that is inspired by the intelligent foraging behavior of honey bees. ABC comprises of three kinds of bees namely (a) Scout (b) Employed and (c) Onlooker bees. The bees intelligently organize themselves and divide the labor to perform the tasks, like searching for the nectar, sharing the information about the food source etc. The position of a food source represents a possible solution to the optimization problem and the nectar amount of a food source corresponds to the quality (fitness) of the associated solution. An epigrammatic mathematical narration of ABC is presented in section 3. Like other population based search heuristics ABC has gained a great attention of researchers and scientists to implement it to solve many real life problems in versatile domain. The latest applications of ABC can be found in [11]-[20]. A brief overview of ABC is given in section 2.

Like other evolutionary algorithms ABC also has some drawbacks which obstruct its performance. ABC is good at exploration while poor at exploitation. Therefore, accelerating convergence speed and avoiding the local optima have become two important and appealing goals in ABC research. A number of ABC variants have, been proposed to achieve these two goals [21]-[27]. The comprehensive survey of ABC can be found in [28]. In the present study we improved the movement of onlooker bees using local search method called golden section search method in order to balance exploration & exploitation and to get more efficient food locations. The proposed *ILS*-ABC is tested four engineering design problems which show its performance over basic ABC.

The rest of the paper is organized as follows: In section 3, the proposed variant *ILS*-ABC is described. The problem set taken to validate the proposal is mentioned in section 4. Experimental settings, comparison criteria and result analyses are given in section 5. The conclusions are presented in section 6.

## 2. Artificial Bee Colony: A Brief Overview

Division of labor and self organization are the component keys in bee colony. In a self-organizing system, each of the covered units may respond to local stimuli individually and act together to accomplish a global task via division of labor without any centralized regulation. According to [29] positive feedback, negative feedback, multiple interactions and fluctuations are the four characteristics on which self-organization rely. Foraging, nest building, marriage, task selection and navigation are the few tasks that bee swarms performs. Task selection is dependent on the environment and hive, which can be changed adaptively. Foraging is one of the major tasks for the bees. There are three types of bees associated with the foraging task, employed, onlooker and scout bees. Similarly ABC tries to model natural behavior of real honey bees in food foraging. The colony of artificial bees also contains three groups of bees: employed bees which are responsible for exploiting the food sources and pass the information to the onlooker bees, which are waiting in the hive. Onlooker bees chooses the food sources by watching the waggle dance performed by employed bees while scouts explores the food sources randomly based on some external clue [30]. This intelligent foraging mechanism can be better explained using the graphical representation shown in figure 1.

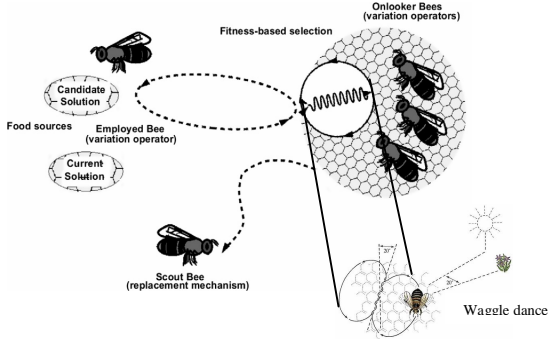

**Figure 1.** Graphical representation of the elements of ABC algorithm

In the initial phase of foraging, bee explores the environment randomly in search of food sources. When the forager bee finds the food source it becomes employed bee. After exploiting the discovered food source, employed bee returns to the hive and unloads the gathered nector. Now it's up to bee to go back directly to the discovered food source or can share the site information to the onlooker bees waiting in the hive by performing a dance called waggle dance on the dance area. Through this dance the employed bee informs the onlooker bee about the direction w.r.t. Sun, distance and quality (Fitness) of the food source. If the food source exploited by employed bee gets exhausted then the bee becomes scout bee, also called the replacement mechanism, and explores the new food sources randomly. The Onlooker bees choose the source site depending on the frequency of the dance, as frequency is proportional to quality of the food source. This is how the onlooker bee becomes the employed bee.

In ABC algorithm the food Sources represents the candidate solutions/possible solutions and the nector amount of the food source represents the fitness associated with the solution for the optimization problem. And each of the food sources is exploited by only one employed bee hence the numbers of food sources are equal to the employed bees.

By this analogy between the intelligent foraging behavior of bees and the ABC algorithm, the basic ABC algorithm can be explained as follows:

*Initialization:*

Like all other population based search heuristics, the first step of ABC is generation of initial population. We assume that the initial population is made up of $SN$ number of $n$-dimensional real-valued vectors, representing the food sources, which are generated randomly. Suppose $X_i = \{x_{i,1}, x_{i,2}, \ldots, x_{i,n}\}$ denotes the $i^{th}$ food source in the population, where each food source is initialized as.

$$x_{i,j} = x_{min,j} + rand(0,1)(x_{max,j} - x_{min,j}) \quad (1)$$

here $i \in 1,2,\ldots,SN$; $j \in 1,2,\ldots,n$. $x_{max,j}$ and $x_{min,j}$ denotes the upper and lower bound constraints respectively of decision variable. These food sources are assigned randomly to $SN$ number of employed bees and their fitness is determined.

*Employed Bees Stage*

The duty of employed bees is to determine new food source say, $v_i$ with the help of the food source $x_i$ assigned to it during the initialization phase. The equation used is:

$$v_{ij} = x_{ij} + \phi_{ij}(x_{ij} - x_{kj}) \quad (2)$$

where $k \in \{1,2,\ldots,SN\}$ and $j \in \{1,2,\ldots,n\}$; Control parameter $\phi_{ij} \in [-1,1]$ gives the step length which decides the movement of bees. $k$ is generated randomly however, $k \neq i$. After a new food source has been generated by the employed bee, a tournament selection is held between $v_i$ and $x_i$ and the one with a better fitness value becomes the member of the population.

*Onlookers Phase*

The work of onlookers is to enhance the exploration capability of ABC. An onlooker evaluates the nectar information (i.e. the fitness of solutions) collected from all the employed bees and selects a food source $x_i$ using probability $P_i$ as:

$$P_i = f_i / \sum_{k=1}^{SN} f_k \quad (3)$$

where $f_i$ denotes the fitness value of the $i^{th}$ food source. The onlooker after selecting the food source $x_i$, modifies it by using Equation (2). Once again a tournament is held between the food source selected and the food source modified by the onlooker and once again as in previous step, the one with better fitness value becomes a member of the population.

*Significance of Scouts*

The main work of scouts is to induce diversity in ABC. An employed bee which is not able to improve the fitness value of a food source becomes a scout. A scout is activated when it is observed that the fitness of a food source $x_i$ shows no betterment even after a specified number of trials limit. Stagnation in the fitness value of a food source indicates that the particular food source may be replaced with a new food source. Scout produces a new food source with the help of equation (1).

**Pseudocode of the ABC:**

*Initialization of the Food Sources*
*Evaluation of the Food Sources*
**Repeat**
*Produce new Food Sources for the employed bees*
*Apply the greedy selection process*
*Calculate the probability values for Onlookers*
*Produce the new Food Sources for the onlookers*
*Apply the greedy selection process*
*Send randomly scout bees*
*Memorize the best solution achieved so far.*
**Until termination criteria are met.**

## 3. Local Search in ABC: *ILS*-ABC

In order to improve the performance of basic ABC, golden section search [31] is introduced in the structure of basic ABC. This process helps in balancing exploration and exploitation. The poor balance may result in weak optimization method which may suffer from trapping in a local optima, stagnation and premature convergence. Introduction of golden section search helps in improving the global convergence and to prevent to stick on a local food source of the ABC. Hence the proposed variant is termed as *ILS*-ABC. In the proposed algorithm the onlooker bee replaces the food sources in basic ABC (Equation 2) using the following Equation:

$$v_{ij} = x_{ij} + F_z(x_{ij} - x_{kj}) \qquad (4)$$

where $F_z$ is scaling factor depends on the value of a & b discussed below. The generation of a new food sources is seen as an opaque procedure depending on the scale factor $F_z$.

Scaling factor becomes an important aspect that is to be controlled in order to guarantee a high quality food source (solution) that can have an important role in the succeeding generations.

A local search procedure is applied to $F_z$ in order to detect the scale factor value which guarantees a food sources with a high performance. This can be gained by minimizing the function $f(F_z)$ in the decision space [−1, 1]. The meaning of the negative scale factor is the inversion of the search direction. If the search in the negative direction succeeds, the corresponding positive value is assigned to the food sources for the subsequent generations.

The Scale factor golden section search applies the Golden section search to the scale factor in order to generate a high quality food sources. This scheme processes in the interval [$a = -1$, $b = 1$] and generates two intermediate points:

$$F_z^1 = b - \frac{b-a}{\delta}; \qquad F_z^2 = a + \frac{b-a}{\delta}$$

where $\delta = \frac{1+\sqrt{5}}{2}$ is the golden ratio. The evaluated values of $f(F_z^1)$ and $f(F_z^2)$ are then compared and if $f(F_z^1) < f(F_z^2)$ then $F_z^2$ replaces b and this procedure is repeated in the new smaller interval [a, b]. The process is described as below:

---
**Repeat**

Compute $F_z^1 = b - \frac{b-a}{\delta}; \qquad F_z^2 = a + \frac{b-a}{\delta}$

Evaluate $f(F_z^1)$ and $f(F_z^2)$

**If** $f(F_z^1) < f(F_z^2)$ then

   b = $F_z^2$

**Else**

   a = $F_z^1$

**End If**

**Until termination criteria are met.**

---

## 4. Test Bed: Real life problems

The performance of *ILS*-ABC is evaluated on test bed of four engineering design real life problems taken from the literature.

The performance of proposed algorithms is evaluated on a test bed of seven unconstrained real life problems that are common in various fields of engineering designs. These are: (A) Optimization of Transistor Modeling (B) Optimal capacity of gas production facility (C) optimal thermohydraulic performance of an artificially roughened air heater and (D) Design of a gear train. All the problems considered in the present study are highly nonlinear in nature. Mathematical models of the problems are discussed below:

**(A) Optimization of transistor modeling [32]:** The objective function of this problem provides a least-sum-of-squares approach to the solution of a set of nine simultaneous nonlinear equations, which arise in the context of transistor modeling. The mathematical model of the transistor design is given by,

*Minimize* $f(x) = \gamma^2 + \sum_{k=1}^{4}(\alpha_k^2 + \beta_k^2)$

Where

$\alpha_k = (1 - x_1 x_2)x_3\{\exp[x_5(g_{1k} - g_{3k}x_7 \times 10^{-3} - g_{5k}x_8 \times 10^{-3})] - 1\}g_{5k} + g_{4k}x_2$

$\beta_k = (1 - x_1 x_2)x_4\{\exp[x_6(g_{1k} - g_{2k} - g_{3k}x_7 \times 10^{-3} + g_{4k}x_9 \times 10^{-3})] - 1\}g_{5k}x_1 + g_{4k}$

$\gamma = x_1 x_3 - x_2 x_4$

*Subject to*: $x_i \geq 0$, i = 1, 2, … , 9

And the numerical constants $g_{ik}$ are given by the matrix

$$\begin{bmatrix} 0.485 & 0.752 & 0.869 & 0.982 \\ 0.369 & 1.254 & 0.703 & 1.455 \\ 5.2095 & 10.0677 & 22.9274 & 20.2153 \\ 23.3037 & 101.779 & 111.461 & 191.267 \\ 28.5132 & 111.8467 & 134.3884 & 211.4823 \end{bmatrix}$$

**(B) Optimal capacity of gas production facilities [33]:** This is the problem of determining the optimum capacity of production facilities that combine to make an oxygen producing and storing system. The mathematical model of this problem is given by,

*Minimize* $f(x) = 61.8 + 5.72x_1 + 0.2623[(40 - x_1)\ln(\frac{x_2}{200})]^{-0.85} + 0.087(40 - x_1)\ln(\frac{x_2}{200}) + 700.23x_2^{-0.75}$

*Subject to:* $x_1 \geq 17.5$, $x_2 \geq 200$

*Bounds:* $17.5 \leq x_1 \leq 40$, $300 \leq x_2 \leq 600$.

**(C) Optimal thermohydraulic performance of an artificially roughened air heater [34]:** In this problem the optimal thermohydraulic performance of an artificially roughened solar air heater is considered. Optimization of the roughness and flow parameters (p/e, e/D, Re) is considered to maximize the heat transfer while keeping the friction losses to be minimum. The mathematical model of this problem is given by,

*Maximize* $f(x) = 2.51 * \ln e^+ + 5.5 - 0.1R_M - G_H$

Where $R_M = 0.95 x_2^{0.53}$; $GH = 4.5(e^+)^{0.28}(0.7)^{0.57}$,
$e^+ = x_1 x_3 (\bar{f}/2)^{1/2}$; $\bar{f} = (f_s + f_r)/2$; $f_s = 0.079 x_3^{-0.25}$;
$f_r = 2(0.95 x_3^{0.53} + 2.5 * \ln(1/2 x_1)^2 - 3.75)^{-2}$;

Subject to: $0.02 \leq x_1 \leq 0.8$, $10 \leq x_2 \leq 40$, $3000 \leq x_3 \leq 20000$

**(D) Design of gear train [32]:** This problem is to optimize the gear ratio for the compound gear train. It is to be designed such that the gear ratio is as close as possible to 1/6.931. For each gear the number of teeth must be between 12 and 60. Since the number of teeth is to be an integer, the variables must be integers. The mathematical model of gear train design is given by,

$$\text{Minimize} \quad f(x) = \left\{ \frac{1}{6.931} - \frac{T_d T_b}{T_a T_f} \right\}^2 = \left\{ \frac{1}{6.931} - \frac{x_1 x_2}{x_3 x_4} \right\}^2$$

Subject to: $12 \leq x_i \leq 60$ $i = 1,2,3,4$

$[x_1, x_2, x_3, x_4] = [T_d, T_b, T_a, T_f]$, $x_i$'s should be integers. $T_a$, $T_b$, $T_d$, and $T_f$ are the number of teeth on gears A, B, D and F respectively.

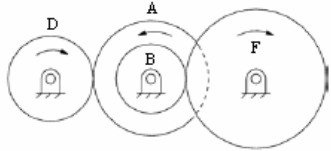

**Figure 2.** Design of Gear Train

## 5. Experimental settings, Result analyses and Comparison criterion

### 5.1. Experimental Settings

ABC and the proposed variant *ILS*-ABC are implemented on Dev-C++ and the experiments are conducted on a computer with 2.00 GHz Intel (R) core (TM) 2 duo CPU and 2-GB of RAM. For each problem, all the algorithms independently run 25 times.

The population size, MCN (maximum cycle numbers) and limit are fixed to 40, 10000 and 100 respectively. And the reported results are the means and standard deviations of the statistical experimental data. The random numbers are generated using inbuilt *rand* () function with same seed for every algorithm.

For each algorithm, the stopping criteria is to terminate the search process when one of the following conditions is satisfied: (i) the maximum number of generations is reached (assumed 1000 generations), (ii) $|f_{max} - f_{min}| < 10^{-4}$ where $f(x)$ is the value of objective function.

### 5.2. Result Analyses and Comparison Criteria

The performances of proposed *ILS*-ABC is demonstrated on a set of four nonlinear engineering design problems and the numerical results are compared with the basic PSO, DE, and ABC. Standard performance measures like average fitness function values, average number of function evaluations and time taken to simulate are calculated. Average fitness function value implies the mean of the best fitness values obtained by the algorithm till the stopping criteria is satisfied, Number of function evaluations (NFE) tells about the speed of an algorithm. Smaller NFE shows that the algorithm converges more quickly. Numerical results based on these performance measures are given in Table 1 and Table 2. From Table 1, which gives the average fitness function value, NFE and execution time, we see that in terms of average fitness function value and time all the algorithms gave more or less similar results although in some cases the proposed algorithms gave a marginally better performance than PSO, DE and ABC. However if we compare the NFE in Table 2 the superior performance of the proposed algorithms become more evident.

Acceleration rate (AR) [35] is used to compare the convergence speeds between *ILS*-ABC and other algorithms. It is defined as follows:

$$AR = \frac{NFE_{one\,algorithm} - NFE_{other\,algorithm}}{NFE_{one\,algorithm}} \%$$

From Table 2 we can see that the proposed *ILS*-ABC gives the better results for every problem in the comparison to the PSO, DE and ABC. Further from the Table 2 it is clearly analysed that the proposed *ILS*-ABC is faster than PSO by 38.6%, DE by 27.7% and faster than ABC by 31.6%.

**Table 1.** Simulation results of four engineering design real life problems

| Item | PSO | DE | ABC | *ILS*-ABC | Source Result |
|---|---|---|---|---|---|
| **(A) *Optimization of Transistor Modeling*** | | | | | |
| $x_1$ | 0.9010 | 0.9013 | 0.9009 | 0.9002 | 0.90 |
| $x_2$ | 0.8841 | 0.5878 | 0.5224 | 0.4519 | 0.45 |
| $x_3$ | 4.0386 | 3.7812 | 1.0764 | 1.0352 | 1.0 |
| $x_4$ | 4.1488 | 3.9021 | 1.9494 | 2.045 | 2.0 |
| $x_5$ | 5.2436 | 5.1962 | 7.8536 | 5.495 | 8.0 |
| $x_6$ | 9.9326 | 11.2697 | 8.8364 | 9.734 | 8.0 |
| $x_7$ | 0.1009 | 0.0979 | 4.7712 | 0.103 | 5.0 |
| $x_8$ | 1.0599 | 1.1053 | 1.0074 | 1.012 | 1.0 |
| $x_9$ | 0.8066 | 0.6799 | 1.8545 | 0.610 | 2.0 |
| $f(x)$ | 0.0695 | 0.0618 | 0.0113 | 0.0304 | NA |
| NFE | 22195 | 19784 | 17901 | 14932 | NA |
| Time | 0.863 | 0.415 | 0.401 | 0.345 | NA |
| **(B) *Optimal Capacity of Gas Production Facilities*** | | | | | |
| $x_1$ | 17.5 | 17.5 | 17.5 | 17.5 | 17.5 |
| $x_2$ | 600 | 593 | 600 | 600 | 465 |
| $f(x)$ | 169.844 | 169.996 | 169.012 | 169.731 | 173.76 |
| NFE | 342 | 324 | 319 | 264 | NA |
| Time | 0.02 | 0.01 | 0.02 | 0.01 | NA |
| **(C) *Optimal Thermo hydraulic Performance of an Artificially Roughened Air Heater*** | | | | | |
| $x_1$ | 0.05809 | 0.18005 | 0.029589 | 0.07111 | 0.052 |
| $x_2$ | 10 | 10 | 9.012 | 9.876 | 10 |
| $x_3$ | 10400.2 | 4918 | 5028.4 | 4220 | 10258 |
| $f(x)$ | 4.2142 | 4.0923 | 4.2166 | 4.0703 | 4.182 |
| NFE | 6207 | 5190 | 4425 | 3026 | NA |
| Time | 0.3 | 0.3 | 0.3 | 0.2 | NA |
| **(D) *Design of Gear Train*** | | | | | |
| $x_1$ | 13 | 16 | 19 | 18.75 | 18 |
| $x_2$ | 31 | 19 | 16 | 13.6 | 22 |
| $x_3$ | 57 | 49 | 44 | 44 | 45 |
| $x_4$ | 49 | 51 | 49 | 52 | 60 |
| $f(x)$ | 9.98e-11 | 2.78e-11 | 2.78e-11 | 5.59e-07 | 5.7e-06 |
| Gear ratio | 0.14429 | 0.14428 | 0.14428 | 0.14424 | 0.14666 |
| Error (%) | 0.007398 | 0.000497 | 0.000447 | 0.000401 | 1.65 |
| NFE | 480 | 340 | 648 | 252 | NA |
| Time | 0.1 | 0.01 | 0.02 | 0.01 | NA |

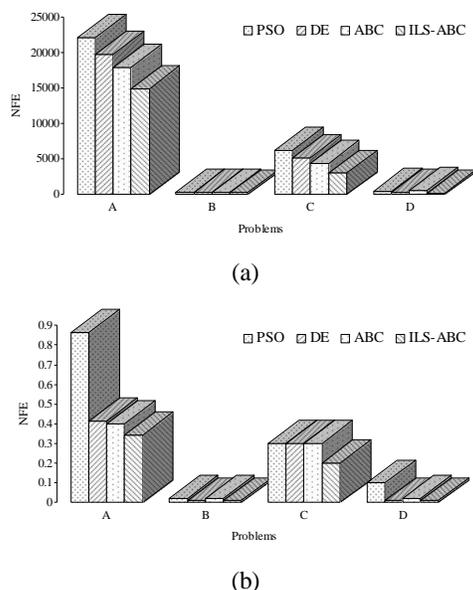

**Figure 1.** **(a)** NFE and **(b)** time taken to compile the four (*A*, *B*, *C* and *D*) engineering design real life problems.

**Table 2.** Comparison of *ILS*-ABC with PSO, DE and ABC in terms of Acceleration rate (%)

| Problems | PSO Vs *ILS*-ABC | DE Vs *ILS*-ABC | ABC Vs *ILS*-ABC |
|---|---|---|---|
| A | 32.7 | 24.5 | 16.6 |
| B | 22.8 | 18.5 | 17.2 |
| C | 51.2 | 41.7 | 31.6 |
| D | 47.5 | 25.9 | 61.1 |
| Average AR (%) | 38.6 | 27.7 | 31.6 |

## 6. Conclusions

In this paper we present an algorithm *ILS*-ABC based on local search mechanism. In *ILS*-ABC golden section method is introduced in the onlooker phase of basic ABC, to improve the local search ability as well as tried to balance the two common and antagonist aspects, exploration and exploitation. The proposed algorithm is applied to four real life engineering design problems. And the simulated results in terms of acceleration rate that directly indicates the better convergence speed, prove the efficiency and applicability of the proposed work.

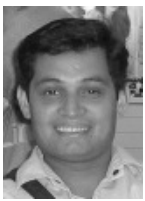

**Tarun Kumar Sharma** did his MCA in 2001, M.Tech (IT) in 2009 and presently pursuing Ph.D from Indian Institute of Technology (IIT) Roorkee, India. He has almost 9 years of teaching experience in Engineering College. His key areas are Evolutionary Computing; Software Engineering; Computer based Optimization Techniques; ERP. His research interest includes swarm intelligence algorithms and their applications in various complex engineering design problems. His publications are in Journals and International Conferences of repute. He volunteered in SocPros-2011, an First International Conference on Soft Computing for Problem solving. He is peer reviewer of many IEEE conferences and International Journals. He is student member of Machine Intelligence Research (MIR) Labs, WA, USA.

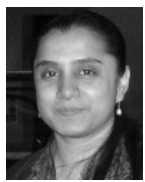

**Millie Pant** is working as an assistant professor in Department of Paper Technology, Indian Institute of Technology (IIT), Roorkee, India since 2007. Her research interest includes evolutionary and swarm intelligence algorithms and their applications in various complex engineering design problems. Her publications are in Journals and International Conferences of repute. She has published over 100 referred on evolutionary algorithms (GA, PSO, DE and ABC) and their applications in electrical design problems, image processing papers. She has been program committee member of over 10 International events and Program Committee Chair of SoCProS-211. She is Program Committee Chair of the 7th International Conference on Bio-Inspired Computing: Theories and Application (BIC-TA 2012) and SoCProS-2012 (International Conference on Soft Computing for Problem Solving).

**V.P. Singh** is Director, SCET Saharanpur, and retired Professor from the department of Paper Technology, IIT Roorkee. His research area includes Mathematical Simulation and Modeling. He has number of publications in journal of repute.